\documentclass[conference]{IEEEtran}

\usepackage{cite}
\usepackage{amsmath,amssymb,amsfonts}
\usepackage{algorithmic}
\usepackage{graphicx}
\usepackage{textcomp,multirow,multicol}
\usepackage{soul,xcolor}
\usepackage{todonotes}

\def\BibTeX{{\rm B\kern-.05em{\sc i\kern-.025em b}\kern-.08em
    T\kern-.1667em\lower.7ex\hbox{E}\kern-.125emX}}

\usepackage{listings}
\usepackage{tikz}
\usetikzlibrary{arrows}
\usetikzlibrary{calc}
\usetikzlibrary{positioning}
\usetikzlibrary{fit}
\usetikzlibrary{quotes}

\usepackage[T1]{fontenc}

\usepackage{url}
\usepackage{subcaption}
\usepackage{graphicx}
\usepackage{booktabs}

\begin{document}

\title{Dynamic Resource Allocation for \\Ensemble Determinization MCTS
\thanks{This article is based on the work of COST Action CA22137 -- ROAR-NET, supported by COST (European Cooperation in Science and Technology).
\\
\indent This research was supported in part by the National Science Centre, Poland, under project number 2021/41/B/ST6/03691 (Jakub Kowalski). 
\\
\indent This work was carried out in part using resources provided by the Wrocław Centre for Networking and Supercomputing (\texttt{http://wcss.pl/}).
} 
}

\makeatletter
\newcommand{\linebreakand}{%
  \end{@IEEEauthorhalign}
  \hfill\mbox{}\par
  \mbox{}\hfill\begin{@IEEEauthorhalign}
}
\makeatother

\author{\IEEEauthorblockN{Jakub Kowalski\IEEEauthorrefmark{1}, Adam Ciężkowski\IEEEauthorrefmark{1}, 
Artur Krzyżyński\IEEEauthorrefmark{1},
Mark H. M. Winands\IEEEauthorrefmark{2}}
\IEEEauthorblockA{\IEEEauthorrefmark{1}\textit{Faculty of Mathematics and Computer Science, University of Wroc{\l}aw}\\jakub.kowalski@cs.uni.wroc.pl, adamciezkowski19@gmail.com,  artur.krzyzynski@gmail.com}
\IEEEauthorrefmark{2}\textit{Department of Advanced Computing Sciences, Maastricht University}\\
m.winands@maastrichtuniversity.nl
}

\IEEEoverridecommandlockouts
\maketitle

\begin{abstract}
Simulation-based algorithms are especially suited for high-uncertainty environments such as adversarial board games with significant elements of randomness and hidden information. 
In particular, several Monte Carlo Tree Search (MCTS) variants are commonly used in such domains.
In this paper, we propose a series of enhancements for Ensemble Determinization MCTS, introducing two axes for dynamic resource allocation. First, Dynamic Number of Determinizations, increases or decreases the number of currently used determinization trees depending on the behavior of so-far search.
Second, Dynamic Simulation Allocation, splits the simulation budget nonuniformly across the determinization trees, using simulation-to-simulation decisions to choose the tree with potentially the best knowledge gain.
As benchmark domains, we used three popular tabletop games: Jaipur, Lost Cities, and Splendor.
Testing our proposed enhancements in iteration- and time-based settings showed that particular configurations yield a statistically significant increase in the algorithm's strength.
\end{abstract}




\section{Introduction} 

In this paper, we investigate enhancements of the Monte Carlo Tree Search (MCTS) algorithm \cite{mctssurvey,swiechowski2023} dedicated to improving its strength in environments with uncertainty. 
Although MCTS has been widely proven to perform extremely well in many deterministic games with perfect information, introducing nondeterminism and hidden information elevates the problem to another level and requires dedicated solutions.
The most common MCTS variants for such domains include Ensemble Determinization MCTS \cite{cowling2012ensemble}, Sample Open Loop MCTS \cite{perez2015open}, and Information Set MCTS \cite{cowling2012ismcts}.

The main contribution of this paper is introducing the concept of dynamic resource allocation for Ensemble Determinization MCTS (ED-MCTS).
We consider Dynamic Number of Determinizations, allowing turn-by-turn adjustment of determinization trees, with adjustable stability, determining how hard it is to convince the algorithm to modify its current behavior.
We also implemented several methods for Dynamic Simulation Allocation, splitting the budget across all determinization trees based on the local contestedness of each tree and the confidence of each move's score.
We tested the proposed enhancements on three popular tabletop games: Jaipur, Lost Cities, and Splendor, in both iteration- and time-based settings.

This paper is structured as follows. 
In the next section, we briefly introduce the MCTS family of search algorithms, with a focus on the variants used for games with imperfect information and randomness.
Section~\ref{sec:dra} describes our proposed dynamic enhancements and the static algorithm variants tested, including final move selection mechanisms, dynamic adjustments to the number of determinizations, and dynamic allocation of simulations.
In Section~\ref{sec:games}, we concisely present board games used as our test domains: Jaipur, Lost Cities, and Splendor.
Section~\ref{sec:experiments} contains the results of our experiments.
Finally, we conclude, identify the limitations of this work, and provide perspective on future research in Section~\ref{sec:conclusions}.


\section{Background}



\subsection{Monte Carlo Tree Search}
Monte Carlo Tree Search \cite{coulom06,Kocsis_2006_Bandit} is a best-first search method that gradually builds up a search tree, balancing exploitation of parts that seem promising based on earlier iterations, with exploration of parts that were infrequently explored. It does this by iterating through four strategic steps (selection, expansion, simulation, backpropagation) until a time or iteration budget expires.
Then, the final move that is ultimately chosen to be played is selected. Usually it is the one from the root node that has the highest visit count, but other strategies are possible \cite{chaslot08}.

In our implementation, we use a simplified AlphaGo-like \cite{silver2016alphago} approach, where the simulation result is substituted by the direct call to the heuristic evaluation function.

The most common MCTS approach uses the UCB1 strategy \cite{auerfinit02} to choose among the children of the current node during the selection. This strategy selects the child achievable by applying an action that satisfies Formula~(\ref{eq:ucb1}):

 \begin{equation}
 \label{eq:ucb1}
 \mathrm{argmax}_{a} \left(Q(n, a) + C \times \sqrt{\frac{\ln{N(n)}}{N(n, a)}}\right),
 \end{equation}

where $Q(n, a)$ is the average reward of action $a$ in tree node $n$, $N(n)$ is the visit count of $n$, and $N(n, a)$ is the number of times action $a$ was selected from state $n$. $C$ is a hyperparameter that can be tuned experimentally. Here, ties are broken randomly.

MCTS has been successfully applied to a number of games, including Go \cite{gelly2011monte}, Lines of Action \cite{winands2010monte}, Hex \cite{arneson2010monte}, 
and also within the General Game Playing domain \cite{finnsson2008simulation,finnsson2010learning,Kowalski2022SplitMoves}

\subsection{Sample Open Loop MCTS}

The open loop approach \cite{perez2015open} is one solution for handling nondeterministic games. It is applicable to a range of simulation-based search algorithms, including the Rolling Horizon Evolutionary Algorithm \cite{perez2013rolling} and Monte Carlo Tree Search. 

In an open-loop approach, every node represents the set of game states that can be reached by playing the sequence of actions corresponding to the path from the root to this state. Thus, by not storing the game states directly in the search tree, the simulations naturally converge to the proper expected values of applying action sequences. 
For the games without imperfect information, the root node represents a single true game state. Otherwise, at the beginning of the MCTS iteration, the actual content of the root node needs to be randomized as well.
In particular, Sample Open Loop MCTS has been used within the General Video Game AI domain \cite{perez20152014,soemers2016enhancements}.

\subsection{Ensemble Determinization MCTS}

The ensemble determinization is an alternative approach that allows the use of a deterministic algorithm in environments with uncertainty. 
The principle behind determinization is that, at the start of each iteration, the hidden information consistent with the history of the game is filled in, and the random choices in the environment are determinized. Thus, it can be seen as instantiating a game simulation model with a specific seed.
Usually, multiple determinizations are used at once in the form of an ensemble, and the final decision is based on all individual preferences.

This approach, although burdened with several shortcomings \cite{frank1998search}, works surprisingly well when applied to practical problems. 
These include domains such as Skat \cite{long2010understanding}, Dou Di Zhu \cite{whitehouse2011determinization},
Magic: The Gathering \cite{cowling2012ensemble}, or Scotland Yard \cite{nijssen2012monte}.

\section{Dynamic Resource Allocation}
\label{sec:dra}

\subsection{Final Move Selection Mechanisms}\label{sec:finmove}

First, let us discuss the potential move selection mechanisms. This is not a ``dynamic'' adjustment, but it is a necessary and impactful step before any other enhancements can be introduced. There are multiple approaches to aggregate the result from a set of determinization trees. 

\begin{itemize}
    \item \emph{visits} -- selecting the move with the highest sum of number of visits in all trees (which is equivalent to the \emph{averaging visits} method from the literature, but simpler).
    \item \emph{score} -- selecting the move with the highest sum of scores in all trees.
    \item \emph{voting} -- when each tree casts a vote on its top-performing move, and the move with the most votes is selected.
    \item \emph{borda} -- selecting the move with the most points, where, within each tree, the move receives one point for every move it outranks in that tree.
\end{itemize}

Best moves within the particular trees, unless otherwise specified, are determined by the number of visits.
Draws in systems other than \emph{visits} are resolved using the \emph{score} system.

\subsection{Dynamic Number of Determinizations}
\label{sec:dyndet}

The standard approach for using Ensemble MCTS is to treat the number of determinization trees as a tunable constant. However, depending on the game rules, the opponent, or the actual line of play, the amount of uncertainty about the game state may vary. Thus, we propose a formula to dynamically allocate the number of such trees depending on the behavior of so-far search.

Our proposed approach and reasoning behind it go as follows. 
If there is a small margin between the winning move and the runner-up, the result is potentially not stable, and to improve the confidence, more determinizations should be used. Thus, we say that \emph{if during the last $L_{\uparrow}$ turns, the winning move achieved a margin that could be modified by changing the outputs of at most $M_{\uparrow}$ trees, in the subsequent turn increase the number of determinization trees by 1.}

Analogously, the number of determinization trees may be considered stable if removing some of them does not change the voting result. This approach can be generalized and expressed by defining that \emph{if during the last $L_{\downarrow}$ turns, the winning move achieved a margin that could not be modified by removing the outputs of at most $M_{\downarrow}$ trees, in the subsequent turn decrease the number of determinization trees by 1.}

Note that these rules are naturally interpretable in all final move selection mechanisms considered in this paper. For example, with $M_{\uparrow}=2$ and a \emph{voting} system, each turn the winning move needs to gather at least 2 more votes above the runner-up to consider this turn stable. 
With $M_{\downarrow}=1$ and a \emph{visits} system, to consider a turn as unstable, the winner's visits must be greater than the runner-up's visits by less than the visits of one of the trees contributing to the winning move.




\subsection{Dynamic Simulation Allocation}
\label{sec:dynamicsims}

As the budget is given each turn for the entirety of the search to perform, it may be beneficial to allocate it nonuniformly across the determinization trees. The most straightforward justification is when in some tree there is a clearly winning move, and the remaining simulations can be put to improve the quality of some other, less unequivocal tree, instead.
Note that the consequences of non-uniform spread of simulations strongly depend on the best move selection mechanisms.


This budget-allocation problem is related to the best arm identification for multi-armed bandits \cite{stephenson2026best,gabillon2011multi}, minimizing simple regret. However, our proposed approaches are strictly tailored to the current use case. Attempts to apply other similar methods are deliberately left as future work.

Our baseline is the naive uniform per-tree allocation that splits the total budget $B = T\cdot N$, assigning $N$ simulations to each of the $T$ determinization trees. Below, we describe a set of policies we proposed and tested, focusing on truly dynamic iteration-by-iteration allocation via informed decisions


To help assess how much each tree needs a better evaluation, we introduce a \emph{local contestedness measure}: for a given tree, the best root child is compared with its strongest alternatives using one-sided confidence bounds (a Hoeffding radius at a configurable confidence level). A tree is \emph{locked} when the lower bound of its best move already exceeds the upper bound of every alternative, meaning its recommendation should no longer change; otherwise, its degree of contention is measured by the overlap between these bounds. 





\subsubsection{Greedy} 
The iterations are given to the currently most-contested unlocked tree. As long as any tree is contested, trees whose top-two root moves have non-overlapping confidence intervals receive no further iterations.

\subsubsection{UCB: win rate difference}

A ``high-level'' UCB-like formula, applied to determinization trees. Thus, for each iteration, we decide which tree it should be allocated to according to

 \begin{equation}
 \label{eq:ucballoc}
 \mathrm{argmax}_{t_i} \left(S(t_i) + C' \times \sqrt{\frac{\ln{N(t)}}{N(t_i)}}\right),
 \end{equation}

where:
\begin{itemize}
\item $C'$ is the exploration constant
\item $N(t_i)$ is the visit count of determinization tree $t_i$,
\item $N(t)$ is the visit count of all trees,
\item $S(t_i)$ is the estimated tree score of the tree $t_i$.
\end{itemize}

The above schema can use different interpretations of the function $S$.
Here, the tree score is $1-(v_1-v_2)$, where $v_1$ and $v_2$ are the mean scores of the best and second-best root moves. 
The intuition behind this choice is that we aim to exploit trees with uncertain results, so the ones where the difference between top moves is the smallest.


\subsubsection{UCB: visits proportion}

The same high-level UCB scheme, with a tree score function $S$ based on visit counts rather than win rates. If $n_1$ and $n_2$ are the visit counts of the most-visited and second-most-visited root moves in a tree, the tree score is $n_2/n_1$. Values close to $1$ indicate that the tree has not yet clearly focused on a single move.





\subsubsection{Across-tree UCB}

Instead of allocating per tree, the ensemble is treated as a single bandit over root moves: per-action statistics are summed across all determinizations, the move maximizing an upper-confidence value is selected, and that move is forced in every tree. This focuses the whole ensemble on the moves that are jointly most promising or most uncertain. Note that a single iteration of this approach can use up to $T$ iterations from the budget.

\subsubsection{Move Pruning}

Root moves whose score upper confidence bound falls below the
lower confidence bound of the current best move are no longer selected inside that tree.
Once a move is eliminated, the iterations that could have tested it are redirected to the
remaining root moves and their subtrees. The \emph{freeze} flag decides how far this may go:
without it, elimination always leaves at least two live root moves, so the tree keeps
searching its leader against a contender; with it, a tree whose leader would eliminate every
rival is declared decided, stops receiving iterations, and the budget is spent on the still-contested trees. These two settings are reported as \emph{Move Pruning} and \emph{Move Pruning + Freeze}.

\section{Testing Domains}
\label{sec:games}

We selected three popular board games as test domains, all of which rely heavily on uncertainty as a key strategic element: Jaipur, Lost Cities, and Splendor.
For each chosen game, the amount and nature of randomness and hidden information differ, allowing us to test our proposed approaches in a variety of circumstances.

One design choice that distinguishes the implementation in this paper from classical MCTS is the replacement of random rollouts with a direct heuristic evaluation function. This makes individual iterations faster and reduces dependence on the quality of random play, at the cost of requiring a reasonably accurate evaluation function for each game. 
The heuristic evaluation functions used for the following experiments are a result of a trial-and-error approach, combining handcrafted ideas with simulation-based parameter search.

\subsection{Jaipur}

Jaipur \cite{pauchon2009jaipur} is a two-player card game designed by Sébastien Pauchon that combines stochastic events with imperfect information. Players take turns collecting camel and goods cards from a shared market and selling sets of goods to earn points. It is described as a fast-paced card game, a blend of tactics, risk, and luck.

The game involves several sources of uncertainty. The order of cards in the deck is randomized, as well as the order of rewards for the bonus token stacks. Each player has only partial knowledge about the opponent's hand, excluding the information about its original content.

From the perspective of game-playing AI, Jaipur offers an interesting balance between strategic depth and computational complexity, combining short tactical decisions with long-term timing considerations. Each turn presents multiple legal actions, including taking cards from the market, exchanging cards, or selling different combinations of goods -- which can produce the order of hundred legal moves. The value of an action often depends on hidden information, future card draws, and the opponent's likely strategy, making accurate evaluation nontrivial.

\subsection{Lost Cities}
Lost Cities \cite{knizia1999lostcities} is a two-player card game designed by Reiner Knizia in which players embark on expeditions to five different destinations represented by colored suits. During each turn, a player either plays a card to an expedition or discards it, and then draws a replacement card. The objective is to maximize the total expedition score by carefully balancing the risks and rewards of investing in different expeditions.

The game combines hidden information with a large amount of stochasticity. Each player's hand is private, while the order of the draw deck is randomized and unknown to both players. As a result, players must estimate the likelihood of drawing useful cards, infer the opponent's intentions from their actions, and decide whether to commit to an expedition before knowing whether sufficient supporting cards will become available.

From the perspective of game-playing AI, Lost Cities presents a challenging sequential decision-making problem. Every move requires trading immediate opportunities against long-term planning, since prematurely starting an expedition may lead to substantial losses if it cannot be completed efficiently. Conversely, delaying an expedition increases the risk that holding cards will reduce our play options in other expeditions or the deck runs out before the expedition's cost is repaid. The relatively small action space is therefore accompanied by a high degree of strategic uncertainty.
This makes Lost Cities a useful contrast to Jaipur: both games contain hidden cards and a draw deck, but they reward different search profiles.

\subsection{Splendor}

Splendor \cite{andre2014splendor} is a two-player strategy board game designed by Marc André in which players compete to accumulate prestige points by acquiring development cards and attracting noble tiles. On each turn, a player may collect gem tokens, reserve a development card, or purchase a card by spending the required resources. Development cards provide both prestige points and permanent gem bonuses that reduce the cost of future purchases, creating an engine-building progression throughout the game.

Splendor has less hidden information than Jaipur and Lost Cities, but it still contains random elements.
The game features stochasticity through the randomized order of development cards in the three decks. While the available cards are visible to both players, the composition of future cards is unknown until they are revealed, requiring players to plan under uncertainty. In contrast to many card games, Splendor is a game of almost perfect information with respect to the current state, as both players see the board, the resources, and every revealed card. The sole exception is a card reserved blindly from a deck, which only its owner may see.

From the perspective of game-playing AI, Splendor presents a challenging optimization problem with a relatively large branching factor. Players must balance immediate gains against long-term engine development, determine which cards to prioritize, and anticipate the opponent's intentions when competing for valuable cards and noble tiles. Decisions often involve delayed rewards, as investing resources in engine-building may temporarily sacrifice scoring opportunities but enable more efficient actions later in the game.

\section{Experiments}
\label{sec:experiments}

The experiments in this section evaluate ED-MCTS performance in Jaipur, Lost Cities, and Splendor. The comparison is mainly framed around a fixed simulation budget per move, as the main research question is how that budget should be allocated. The final experiment additionally repeats a selection of configurations under a fixed wall-clock budget. The uncertainty values used, are approximate $95\%$ confidence half-widths for the reported score. All reported scores belong to an ED-MCTS variant, whose opponent is the Sample Open Loop MCTS, as using SOLMCTS as a baseline is a common practice for comparing tree search algorithms in nondeterministic environments \cite{perez2015open,soemers2016enhancements}.

\subsection{Move Selection Mechanism}

First, as a preliminary but necessary experiment, we will compare the methods of final move selection described in Section~\ref{sec:finmove}.
We test the number of determinization trees and the final move selection rule while keeping the total simulation budget fixed. The case $T=1$ is a single determinized tree, so all aggregation rules collapse to the same decision. For $T>1$, the experiment measures whether the additional breadth over determinizations compensates for the reduced number of iterations available inside each tree. The results are presented in Table~\ref{tab:finmove}.

\begin{table}[ht]\renewcommand{\arraystretch}{1.1}
\caption{The results of ED-MCTS (with uniform per-tree simulation allocation) versus SOLMCTS, comparing final move selection mechanisms over a range of determinization trees. Best value for each value of $T$ in bold. (2000 games, 1000 symmetric pairs, 250k simulations per turn for both agents).}
\label{tab:finmove}
\begin{center}\begin{tabular}{l||c|c|c|c}
\toprule
T & \emph{visits} & \emph{score} & \emph{voting} & \emph{borda} \\ \hline

\multicolumn{5}{c}{Jaipur} \\
$T=1$ & \multicolumn{4}{c}{$\textbf{40.6}\pm2.16$ } \\
$T=2$ & $47.1\pm2.19$ & $\textbf{48.1}\pm2.19$ & $42.7\pm2.17$ & $40.1\pm2.15$ \\
$T=5$ & $51.6\pm2.19$ & $\textbf{53.8}\pm2.18$ & $47.6\pm2.19$ & $45.5\pm2.19$ \\
$T=10$ & $\textbf{53.4}\pm2.19$ & $50.5\pm2.19$ & $49.0\pm2.19$ & $48.2\pm2.19$ \\
$T=25$ & $46.9\pm2.19$ & $\textbf{48.9}\pm2.19$ & $47.8\pm2.19$ & $43.4\pm2.17$ \\
\hline

\multicolumn{5}{c}{Lost Cities} \\
$T=1$ & \multicolumn{4}{c}{$\textbf{40.9}\pm2.15$} \\
$T=2$ & $\textbf{47.9}\pm2.18$ & $46.8\pm2.18$ & $44.6\pm2.17$ & $45.3\pm2.18$ \\
$T=5$ & $\textbf{50.5}\pm2.19$ & $50.1\pm2.18$ & $50.4\pm2.19$ & $48.6\pm2.18$ \\
$T=10$ & $\textbf{50.2}\pm2.18$ & $49.6\pm2.18$ & $49.0\pm2.19$ & $47.8\pm2.19$ \\
$T=25$ & $44.0\pm2.17$ & $\textbf{45.1}\pm2.17$ & $44.2\pm2.18$ & $43.1\pm2.16$ \\
\hline

\multicolumn{5}{c}{Splendor} \\
$T=1$ & \multicolumn{4}{c}{ $\textbf{52.0}\pm2.18$} \\
$T=2$ & $63.7\pm2.11$ & $\textbf{64.7}\pm2.09$ & $64.3\pm2.09$ & $61.7\pm2.13$ \\
$T=5$ & $\textbf{64.4}\pm2.10$ & $63.9\pm2.10$ & $62.5\pm2.11$ & $60.3\pm2.14$ \\
$T=10$ & $59.8\pm2.15$ & $\textbf{60.4}\pm2.15$ & $59.5\pm2.15$ & $58.2\pm2.16$ \\
$T=25$ & $\textbf{52.3}\pm2.19$ & $\textbf{52.3}\pm2.19$ & $50.0\pm2.18$ & $51.6\pm2.18$ \\
\hline

\bottomrule
\end{tabular}
\end{center}
\end{table}

The results show a clear non-monotonic depth--breadth trade-off. A single determinized tree performs worse than the multi-tree version for all games. The win rates achievable with ensemble determinizations are able to significantly improve over this baseline: +13.2 percentage points for Jaipur, +9.6 for Lost Cities, and +12.7 for Splendor. For Jaipur and Splendor, multiple ED-MCTS variants achieve confident wins over the SOLMCTS, while for Lost Cities the best results are close to parity.

Generally, the best results are obtained for a moderate number of determinization trees (5--10), but detailed numbers also depend on the final move selection method. Our results here are surprising, as they do not support the common conclusion from previous publications that majority-based voting is considered superior~\cite{cowling2012ensemble,nijssen2012monte}. Based on our experiments, the \emph{voting} approach is usually within the confidence interval range with dominating \emph{visits}/\emph{score}, but not better. The \emph{borda} approach, at least in the presented formulation, falls behind other selection mechanisms.

\subsection{Dynamic Number of Determinizations Trees}
\label{sec:dynamic-trees}
The next experiment tests whether it is beneficial to adjust the number of determinization trees between turns, according to the idea proposed in Section~\ref{sec:dyndet}. 

The mechanism uses the margin of the ensemble decision after each move to choose the number of trees for the next move. For \emph{voting}, this margin is the vote lead of the winning root action over the runner-up. For summed \emph{visits}, the analogous margin is expressed in tree units: it is the number of winner-supporting trees whose visit contribution would have to be removed before the leader could lose its aggregate advantage. A small margin means that the ensemble is contested; a large margin means that the current number of determinizations appears more than sufficient. 

Dynamic rules for increasing and decreasing the number of trees are enabled simultaneously. If the margin is at most $M_{\uparrow}$ for $L_{\uparrow}$ consecutive turns, the next move uses one additional tree. If the margin is at least $M_{\downarrow}$ for $L_{\downarrow}$ consecutive turns, the next move uses one fewer tree. We use a shared window length, i.e., $L_{\uparrow}=L_{\downarrow}$.
The experiment starts from strong static configurations identified in Table~\ref{tab:finmove}, with $T_0$ denoting the initial number of determinization trees.

The results of our experiment are presented in Table~\ref{tab:dyntrees}, in the form of an ablation study, i.e., as percentage-point win rate differences relative to the corresponding non-dynamic variants presented before. 

\begin{table*}[!htp]\renewcommand{\arraystretch}{1.1}
\caption{The results of ED-MCTS (with uniform per-tree simulation allocation) versus SOLMCTS, testing dynamic number of determinizations. Values for starting number of trees and move selection are given for each batch of experiments. Results are presented as differences with the non-dynamic variant with the same number of initial trees, shown in Table~\ref{tab:finmove}. (1000 games, 500 symmetric pairs, 250k simulations per turn).}
\label{tab:dyntrees}
\begin{center}
\scriptsize
\begin{tabular}{l||c|c|c|c|c|c|c}
\toprule
\multirow{2}{*}{$L_{\uparrow},L_{\downarrow}$ }& $M_{\uparrow}\!=\!1$ & $M_{\uparrow}\!=\!1$ & $M_{\uparrow}\!=\!2$ & $M_{\uparrow}\!=\!2$ & $M_{\uparrow}\!=\!2$ & $M_{\uparrow}\!=\!3$ & $M_{\uparrow}\!=\!3$ \\
& $M_{\downarrow}\!=\!1$ & $M_{\downarrow}\!=\!2$ & $M_{\downarrow}\!=\!1$ & $M_{\downarrow}\!=\!2$ & $M_{\downarrow}\!=\!3$ & $M_{\downarrow}\!=\!2$ & $M_{\downarrow}\!=\!3$ \\
\hline

\multicolumn{8}{c}{Jaipur (\emph{voting}, $T_0=10$)} \\
$1$ & $+1.2$ & $+1.5$ & $+0.9$ & $\textbf{+3.3}$ & $+0.2$ & $+0.7$ & $+2.4$ \\
$2$ & $+1.7$ & $+1.9$ & $-1.1$ & $+1.8$ & $-2.4$ & $-2.1$ & $+3.1$ \\
$3$ & $+2.1$ & $+0.2$ & $-1.7$ & $+0.3$ & $0.0$ & $+0.1$ & $+2.3$ \\
\hline

\multicolumn{8}{c}{Jaipur (\emph{visits}, $T_0=10$)} \\
$1$ & $-1.2$ & $-2.5$ & $-2.7$ & $-1.0$ & $-2.7$ & $\textbf{+2.3}$ & $-3.9$ \\
$2$ & $+1.8$ & $-1.2$ & $+1.6$ & $-1.9$ & $-1.8$ & $-1.3$ & $+2.2$ \\
$3$ & $+1.3$ & $-1.9$ & $-2.3$ & $-0.8$ & $-2.4$ & $-0.8$ & $-1.6$ \\
\hline
\multicolumn{8}{c}{Jaipur (\emph{score}, $T_0=5$)} \\
$1$ & $-4.2$ & $-2.4$ & $-0.8$ & $+0.5$ & $-0.9$ & $-1.7$ & $-2.7$ \\
$2$ & $-1.9$ & $-1.5$ & $-1.2$ & $-2.0$ & $+0.3$ & $+0.1$ & $-3.8$ \\
$3$ & $-3.3$ & $-0.1$ & $-3.1$ & $-3.6$ & $-3.5$ & $-0.1$ & $\textbf{+0.6}$ \\
\hline
\multicolumn{8}{c}{Lost Cities (\emph{voting}, $T_0=5$)} \\
$1$ & $-4.8$ & $-1.6$ & $-0.8$ & $+0.1$ & $+0.2$ & $\textbf{+0.3}$ & $-1.7$ \\
$2$ & $-4.0$ & $-1.6$ & $-0.4$ & $-0.7$ & $-0.4$ & $-1.6$ & $-0.7$ \\
$3$ & $-0.4$ & $-0.7$ & $-1.8$ & $-0.7$ & $-0.3$ & $-1.0$ & $-0.1$ \\
\hline
\multicolumn{8}{c}{Lost Cities (\emph{visits}, $T_0=5$)} \\
$1$ & $-2.6$ & $-3.1$ & $-1.7$ & $-1.8$ & $-3.2$ & $-3.8$ & $-6.4$ \\
$2$ & $-0.7$ & $-2.1$ & $-0.6$ & $-2.0$ & $-0.7$ & $-3.0$ & $-3.0$ \\
$3$ & $-2.0$ & $-0.6$ & $+0.7$ & $\textbf{+1.4}$ & $-3.2$ & $+0.9$ & $-1.4$ \\
\hline
\multicolumn{8}{c}{Splendor (\emph{voting}, $T_0=5$)} \\
$1$ & $-1.3$ & $+3.0$ & $-0.8$ & $+0.4$ & $+0.1$ & $+2.2$ & $-1.6$ \\
$2$ & $+0.8$ & $+1.5$ & $+1.2$ & $-1.9$ & $+0.9$ & $+2.4$ & $-2.1$ \\
$3$ & $+1.7$ & $\textbf{+5.1}$ & $+2.1$ & $+4.2$ & $-0.7$ & $-1.6$ & $0.0$ \\
\hline
\multicolumn{8}{c}{Splendor (\emph{visits}, $T_0=5$)} \\
$1$ & $-3.5$ & $-1.4$ & $-1.1$ & $-6.3$ & $-4.0$ & $-7.2$ & $-4.8$ \\
$2$ & $-1.8$ & $-2.6$ & $-1.8$ & $-3.8$ & $-0.3$ & $-1.6$ & $-2.5$ \\
$3$ & $-2.9$ & $-4.4$ & $-3.3$ & $-2.3$ & $-1.7$ & $\textbf{+0.1}$ & $-2.4$ \\
\hline

\bottomrule
\end{tabular}
\end{center}
\end{table*}

The dynamic tree-count results are more mixed than the static sweep. This is expected: changing the number of trees between turns is an indirect adaptation mechanism. It does not improve the current move directly; it uses the decisiveness of previous ensemble decisions as a signal for how much determinization breadth may be useful on the next move. 

The method is able to confidently improve results for Jaipur (+3.3 percentage points) and Splendor (+5.1). Note that both these results were obtained for the \emph{voting} move selection method, which suggests that it is more susceptible to this type of improvement.
The results for Lost Cities are not so positive, with the largest gain (+1.4) obtained in \emph{visits}-based configuration and not being statistically confident; also, the majority of the results for this game are strictly negative.

Overall, dynamic adjustment of the number of determinizations is a useful mechanism to test, with significant potential gains, but its effect is clearly game- and aggregation-dependent. The method is therefore best viewed as an adaptive refinement of ED-MCTS rather than as a universally beneficial replacement for a well-chosen static tree count.

\subsection{Dynamic Budget Allocation}

The next experiment tests whether uneven distribution of iterations across the determinization trees can be beneficial to the overall search results. We test the policies proposed in Section~\ref{sec:dynamicsims} with both summed-\emph{visits} aggregation and \emph{voting}, because the effect of uneven allocation can depend on how the trees are later merged. Under \emph{voting}, each tree contributes one final vote regardless of how many iterations it received. Under \emph{visits}, a tree that receives more iterations can also have more influence on the final aggregate. 

The results presented in Table~\ref{tab:reallocation} report percentage-point win rate differences relative to the matching static configuration in Table~\ref{tab:finmove}, using the same game, tree count, and final move selection rule. Positive values mean that the policy improved the ED-MCTS score against SOLMCTS; negative values mean that it hurt performance. For \textit{Move Pruning}, the elimination confidence is $\delta=0.05$, with a minimum of $10$ visits before a move can be pruned. The \emph{Move Pruning + Freeze} row combines \textit{Move Pruning} with freezing a tree once its root decision is locked, redistributing its remaining budget to the still-contested trees.

\begin{table*}[htb]\renewcommand{\arraystretch}{1.1}
\caption{Percentage-point differences of dynamic budget-allocation and root-pruning policies relative to the matching uniform-allocation baseline from Table~\ref{tab:finmove}. (1000 games, 500 symmetric pairs, 250k simulations per turn for both agents).}
\label{tab:reallocation}
\centering
\begingroup
\scriptsize
\setlength{\tabcolsep}{6pt}
\makebox[\textwidth][c]{%
\begin{tabular}{l||cc|cc|cc|cc|cc|cc}
\toprule
\multirow{3}{*}{Policy} & \multicolumn{4}{c|}{Jaipur} & \multicolumn{4}{c|}{Lost Cities} & \multicolumn{4}{c}{Splendor} \\
 & \multicolumn{2}{c|}{\emph{visits}} & \multicolumn{2}{c|}{\emph{voting}} & \multicolumn{2}{c|}{\emph{visits}} & \multicolumn{2}{c|}{\emph{voting}} &\multicolumn{2}{c|}{\emph{visits}} & \multicolumn{2}{c}{\emph{voting}}  \\
& $T=5$ & $10$ & $5$ & $10$ & $5$ & $10$ & $5$ & $10$ & $5$ & $10$ & $5$ & $10$\\
\hline
\emph{Greedy} & $-8.0$ & $-9.2$ & $+2.8$ & $-0.1$ & $-3.3$ & $-1.1$ & $-0.3$ & $+\textbf{2.5}$ & $+0.3$ & $-1.3$ & $+1.3$ & $-1.6$ \\
\emph{UCB: win rates difference} & $-14.6$ & $-12.5$ & $-1.8$ & $+0.8$ & $-3.5$ & $-3.7$ & $-0.6$ & $+2.1$ & $-4.7$ & $-2.5$ & $+0.6$ & $+2.5$ \\
\emph{UCB: visits proportion} & $-15.7$ & $-16.5$ & $+1.8$ & $+0.8$ & $-11.1$ & $-4.9$ & $+1.5$ & $-0.7$ & $-4.7$ & $-0.3$ & $-1.5$ & $+2.5$ \\
\emph{Across-tree UCB} & $+\textbf{2.6}$ & $-1.6$ & $+\textbf{4.7}$ & $+3.2$ & $-0.3$ & $+0.4$ & $+0.6$ & $-0.5$ & $-3.3$ & $+\textbf{1.6}$ & $+2.2$ & $+\textbf{2.7}$ \\
\emph{Move Pruning} & $-3.4$ & $-2.3$ & $+0.1$ & $+1.9$ & $+1.6$ & $+\textbf{2.0}$ & $-2.7$ & $+1.5$ & $-1.6$ & $-1.0$ & $-1.3$ & $-1.6$ \\
\emph{Move Pruning + Freeze} & $-9.4$ & $-7.8$ & $-1.7$ & $+1.3$ & $-1.4$ & $+0.9$ & $+1.2$ & $+2.0$ & $-4.3$ & $-0.1$ & $+1.3$ & $+1.1$ \\

\bottomrule
\end{tabular}
}
\endgroup
\end{table*}

The results show that dynamic budget allocation is highly sensitive to both the aggregation rule and the game. The clearest positive pattern is obtained by \textit{Across-tree UCB} with \textit{voting}, which improves Jaipur and Splendor for both tested tree counts. In contrast, the \textit{Tree-level UCB} variants are strongly harmful with summed-visits aggregation, especially in Jaipur, where they reduce the score by more than ten percentage points. This suggests that changing the number of iterations per tree interacts badly with visit-sum aggregation, because the final move then gives more weight to the trees that received more budget. \textit{Move Pruning} is mildly positive in some Lost Cities and Jaipur voting settings, but does not produce a consistent improvement across the table.

\subsection{Combined Results and Timed Experiments}

The previous two experiments tested one adaptive mechanism at a time, each measured against a static baseline. The final experiment tests to what extent both methods can be composed together. Also, as so far we have tested only iteration-based comparison between ED-MCTS and SOLMCTS, thus we also want to test how the achieved win rates translate to the time-based scenario. 

The results of this experiment are presented in Table~\ref{tab:timed}. For a given game, move selection mechanism, and number of (initial) trees, we show the win rate against SOLMCTS under simulation- and time-based settings. For the simulation-based setting, we additionally report the difference in percentage points relative to the appropriate baseline and a ``theoretical'' difference, obtained by summarizing the gains from Tables~\ref{tab:dyntrees} and \ref{tab:reallocation}.

\begin{table*}[htb]\renewcommand{\arraystretch}{1.1}
\caption{Comparison of selected ED-MCTS configurations with combined dynamic resource allocation improvements against SOLMCTS under either a fixed budget of 250k simulations per move or a one-second per move time budget. 
The \emph{Diff.} column is the measured difference from the static baseline of the same game and move selection mechanism, while the \emph{Pred.\ Diff.} column is the sum of the improvements measured separately for the selected dynamic-tree and allocation mechanisms in the previous sections.
(1000 games, 500 symmetric pairs)}
\label{tab:timed}
\centering
\scriptsize
\setlength{\tabcolsep}{4pt}
\begin{tabular}{l|p{3.2cm}|p{2.3cm}||c|c|c||c}
\toprule
Move & Dynamic & Simulation &  \multicolumn{3}{c||}{250k simulations} & \multicolumn{1}{c}{1 second}  \\
Selection & Determinizations & Allocation & Score & Diff. & Pred.\ Diff. & Score  \\
\hline

\multicolumn{7}{c}{Jaipur, $T_0 = 10$} \\
\emph{voting} & off & uniform & $49.0$ & $--$ & $--$ & $50.7$ \\
\emph{voting} & $L=1$, $M_{\uparrow}=2$, $M_{\downarrow}=2$ & Across-tree UCB & $51.9$ & $+2.9$ & $+6.5$ & $50.1$ \\
\emph{voting} & $L=1$, $M_{\uparrow}=2$, $M_{\downarrow}=2$ & Move Pruning & $51.5$ & $+2.5$ & $+5.2$ & $50.2$ \\
\emph{visits} & off & uniform & $53.4$ & $--$ & $--$ & $52.4$ \\
\emph{visits} & $L=1$, $M_{\uparrow}=3$, $M_{\downarrow}=2$ & uniform & $\textbf{55.7}$ & $+2.3$ & $+2.3$ & $\textbf{53.0}$ \\
\hline
\multicolumn{7}{c}{Lost Cities, $T_0 = 5$} \\
\emph{voting} & off & uniform & $50.4$ & $--$ & $--$ & $47.6$ \\
\emph{voting} & $L=1$, $M_{\uparrow}=3$, $M_{\downarrow}=2$ & UCB: visits prop. & $\textbf{53.8}$ & $+3.4$ & $+1.8$ & $\textbf{54.6}$ \\
\emph{visits} & off & uniform & $50.5$ & $--$ & $--$ & $51.4$ \\
\emph{visits} & $L=3$, $M_{\uparrow}=2$, $M_{\downarrow}=2$ & Move Pruning & $49.0$ & $-1.5$ & $+3.0$ & $47.6$ \\
\hline
\multicolumn{7}{c}{Splendor, $T_0 = 5$} \\
\emph{voting} & off & uniform & $62.5$ & $--$ & $--$ & $70.9$ \\
\emph{voting} & $L=3$, $M_{\uparrow}=1$, $M_{\downarrow}=2$ & Across-tree UCB & $64.6$ & $+2.1$ & $+7.3$ & $63.1$ \\
\emph{voting} & $L=3$, $M_{\uparrow}=2$, $M_{\downarrow}=2$ & Greedy & $\textbf{64.7}$ & $+2.2$ & $+5.5$ & $64.8$ \\
\emph{visits} & off & uniform & $64.4$ & $--$ & $--$ & $\textbf{72.3}$ \\
\hline

\bottomrule
\end{tabular}
\end{table*}

The fixed-iteration validation shows that the effects of combining two adaptive mechanisms are not additive. In Jaipur, the two combined voting configurations improve on the static voting baseline by $2.9$ and $2.5$ percentage points, substantially less than the additive predictions of $6.5$ and $5.2$ points. In Splendor, the combined gains are about $2.1$--$2.2$ points, again below the predicted $5.5$--$7.3$ points. In particular, it seems that Simulation Allocation can hinder the gains from successful Dynamic Determinization application. However, Lost Cities is the exception: the selected voting configuration gains $3.4$ points, exceeding its additive prediction of $1.8$ points. These discrepancies show that both tested enhancements should be evaluated jointly rather than selected solely from separate sweeps. 

The results of the time-based experiment confirm that ED-MCTS holds up well under a realistic wall-clock budget. Static summed-visits ED-MCTS stays above parity in every game and is markedly the strongest configuration in Splendor, where it reaches $72.3\%$. Adaptation continues to pay off where it is most needed: in Lost Cities the selected voting configuration climbs from $47.6\%$ to $54.6\%$, a $7.0$-point gain that turns a below-parity baseline into a clear win, and in Jaipur the dynamic summed-visits configuration adds a further improvement, from $52.4\%$ to $53.0\%$. The remaining cases show that the ranking of mechanisms can shift between a fixed iteration count and a fixed time limit. The practical lesson is that mechanism selection should be validated under the deployment budget itself, as a configuration tuned per iteration is not guaranteed to win in the timed setting per second.

\section{Conclusions}
\label{sec:conclusions}

The paper introduces the concept of dynamic resource allocation applicable to Ensemble Determinization MCTS. It proposes two possible axes of such research: Dynamic Number of Determinizations and Dynamic Simulation Allocation. In both types of enhancements, we define and test a selection of approaches. 
The series of experiments we conducted on tabletop board games: Jaipur, Lost Cities, and Splendor, assesses each proposed enhancement. According to the results from both fixed-iteration budget and time-based experiments, some of the proposed enhancements are beneficial, confidently improving win rates against Sample Open Loop MCTS.

Several limitations qualify these conclusions. The comparison is confined to three games and primarily to a specific simulation budget. The best ensemble size and aggregation rule are game-specific, so the numbers should be read as evidence for a trade-off rather than as universally optimal settings. Especially taking into account that playing strength also depends on handcrafted game-specific heuristic evaluation functions. 
Extended experiments should cover more granular search over the number of determinization trees and tests of simulation allocation methods combined with score-based final move selection.

In summary, the gains obtained are highly game-dependent, but the overall approach is potentially beneficial. The mentioned limitations will be addressed in follow-up research.










\bibliographystyle{IEEEtran}
\bibliography{bibliography}

\end{document}